# NAVREN-RL: Learning to fly in real environment via end-to-end deep reinforcement learning using monocular images

Malik Aqeel Anwar[1], Arijit Raychowdhury[2]
*Department of Electrical and Computer Engineering*
*Georgia Institute of Technology, Atlanta, GA, USA*
aqeel.anwar@gatech.edu[1], arijit.raychowdhury@ece.gatech.edu[2]

*Abstract*—We present NAVREN-RL, an approach to *NAV*igate an unmanned aerial vehicle in an indoor *R*eal *EN*vironment via end-to-end reinforcement learning (*RL*). A suitable reward function is designed keeping in mind the cost and weight constraints for micro drone with minimum number of sensing modalities. Collection of small number of expert data and knowledge based data aggregation is integrated into the RL process to aid convergence. Experimentation is carried out on a Parrot AR drone in different indoor arenas and the results are compared with other baseline technologies. We demonstrate how the drone successfully avoids obstacles and navigates across different arenas. Video of the drone navigating using the proposed approach can be seen at https://youtu.be/yOTkTHUPNVY

## I. INTRODUCTION

Over the past decade, there has been considerable success in using Unmanned Aerial Vehicles (UAVs) or drones in varied applications such as reconnaissance, surveying, rescuing and mapping. Irrespective of the application, navigating autonomously is one of the key desirable features of UAVs both indoors and outdoors. Several solutions have been proposed to make drones autonomous in an indoor environment. There has been significant work towards using additional dedicated sensing modalities such as RADAR [1] and LIDAR [2], which provide high accuracy in navigation and obstacle avoidance thus enabling autonomous flights possible. But when the payload and the cost is taken into account, such systems are heavy and expensive, making them almost impossible to be used in low cost Micro Aerial Vehicles (MAV). Ultrasonic SONAR is a cheap alternative but suffers from lack of accuracy and reduced field of view (FOV). Hence for MAVs, using the on-board and relatively cheap sensors, in particular cameras, is an attractive option for autonomous indoor navigation.

In recent years, RL has been extensively explored for different type of robotic tasks. The model-free nature of RL makes it suitable in the problems where little or nothing is known about the environment. RL has been successfully implemented in games and has shown beyond human level performance [3], [4]. However, RL is a data-hungry method and often requires more data compared to other supervised techniques to generate comparable results. The requirement of a large training data-size is often addressed by training in a simulated environment. However, if the environment is unknown, off-line training presents low accuracy and higher crash rates. So far, limited success has been achieved training in real environments. Further, ensuring safety of the agent during training is also challenging.

In this paper we explore a single-camera-based autonomous navigation and obstacle avoidance for MAVs in real environments. Traditional systems employ handcrafted sensing and control algorithms to allow navigation and has led to significant progress in this field [5], [6]. Recently, the success of deep neural networks have enticed researchers to study neuromorphic models of autonomous navigation [7]–[9]. In spite of the success of such machine learning models, we also recognize that true autonomy in intelligent agents will only emerge when bio-mimetic systems can perform continuous learning through interactions with the environment.

The main contributions of the paper are as follows:
- Demonstration of end-to-end Deep RL for collision avoidance using monocular images only and without the use of any other sensing modality.
- Overcoming the issues associated with the implementation of RL in real environments by designing a suitable reward function that takes into account both the safety and sensor constraints.
- Using expert data and knowledge based data aggregation to improve the RL convergence in real time.

## II. RELATED WORK

Our principal goal is to enable the UAV to fly by itself in a real environment, without incurring any additional hardware or sensor cost. Most of the low-cost MAVs come equipped with an on-board camera and Inertial Measuring Unit (IMU). So the use of image frames for navigation is an area of active research. We have studied supervised learning for drone navigation. [10] collects a data-set of 11,500 videos of crashing and learns a neural network that classifies an image as "crash" or "no crash". Based on that knowledge, the authors use a handcrafted algorithm to steer and navigate the drone away from obstacles. [11] uses an indoor data-set and classifies the images according to the action taken by the drone. They define a set of five actions in the action space of the drone, hence posing the problem as a classification problem with five classes. A supervised image classifier with three classes is used in [12] to train a deep neural network for forest navigation. The data-set is collected by mounting three cameras on a hiker's head facing forward, left and right. [13] uses RL as the online learning mechanism to navigate a drone in a forest. A camera frame is taken and is pre-processed before it is fed to the RL system. This pre-processing uses handcrafted algorithms to extract lower dimensional features from the camera image.

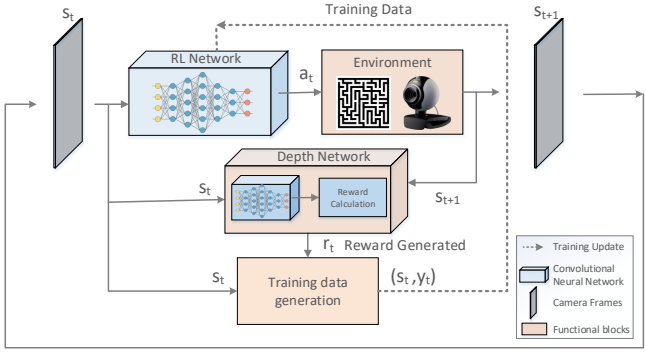

Fig. 1. Block diagram of the key algorithmic components that enable end-to-end RL for obstacle avoidance and autonomous flight in a drone.

[14] uses simulated environments with a larger set of action space (1681 actions). The agent is trained for a deep neural network in 9 simulated environments and the performance is reported. The neural network trained in the simulations is then also tested in the real environment without any fine-tuning and has shown to perform well. Unfortunately, the performance of this approach greatly depends upon the correlation of the simulated and real environment. For the cases of unknown environments which has limited similarity with the simulated training environments, the agent is expected to behave poorly.

All of these previous demonstrations and approaches, in spite of their many successes, either require considerable human/expert intervention, handcrafted algorithms or are implemented offline in simulations, where the simulated and the real endowments need to be nearly identical.

## III. DEEP REINFORCEMENT LEARNING (DRL)

### A. Background on Reinforcement Learning (RL)

The idea of RL is to learn a control policy by interacting with the environment. In this paper, the goal achieved through RL is to take actions that lead to a collision free flight of the drone in a real environment. There is no goal position and the objective is to navigate through the arena safely. Consider the task of obstacle avoidance where the drone interacts with the environment in a sequence of actions, observations and reward calculations. At each time instant, the drone observes the current camera frame $s$. It takes an action $a$ from an action space $\mathcal{A}$ and implements it. Implementing the action moves the drone to a new position where it observes a new camera frame $s'$. This new camera frame along with the action taken will quantify a reward $r$. The goal of the system is to take actions maximizing the long term reward, i.e. at each time step $t$, we need to take an action that eventually leads us to a sequence of states $s_i$ with rewards $r_i$ for $i \in \{t+1, t+2, ...\}$ such that the future discounted return $R_t = \sum_{i=t}^{T} \gamma^{i-t} r_i$ is maximized, where $\gamma \in [0,1]$ is the discount factor. Each of the state-action pair is assigned a Q value $Q(s,a)$. During the learning phase these Q values are updated according to the Bellman optimality equation as follows

$$Q(s,a) = r + \gamma max_{a'} Q(s',a') \qquad (1)$$

Bellman equation update ensures that in a given state $s_t$ selecting an action $a_t = max_{a'} Q(s_t, a')$ will result in maximizing the future discounted reward $R_t$. These Q values are stored as an approximation of a function with states as input. In Deep Reinforcement Learning (DRL) the function to estimate these Q values is a deep neural network.

### B. Challenges of implementing DRL in real environments

RL in real environments for collision avoidance is challenging, as listed below. The methodologies adopted in this paper to address them are described in the next section.

*1) Reward generation:* In real environments, the position of the agent and its distance from obstacles is not known. Hence extra sensing capabilities need to be added to the agent giving it a notion of depth which not only adds to the computation cost but also to the weight of the agent. In this paper, we demonstrate DRL using a single monocular camera.

*2) Safety issues:* RL works via a trial and error method. It is designed to learn from mistakes. For the task of collision avoidance, it means that the agent has to collide into the obstacles to learn. This collision can not only harm the agent, but also the environment. We propose a method of *virtual crash* and a *crash reward* to address this issue.

*3) Resetting the agent to a suitable initial position:* RL requires that the agent must be placed at proper initial position (usually the same) every time it crashes with an obstacle. In simulations, it is trivially achieved while in real environments it poses a challenge. We demonstrate a method of *un-doing the drone's actions* to achieve the same effect as resetting the drone's position.

*4) Large online data-set requirement:* The amount of data required for implementing RL is large. Such training data requirement stems from the fact that the agent starts with little knowledge of the environment and takes random actions to explore it. As opposed to simulations where you can easily collect a large number of data-points, the data-set that can be collected in a real environment is limited. We use several techniques to address this issue, as described in the next section.

## IV. NAVIGATION IN REAL ENVIRONMENTS VIA RL (NAVREN-RL)

We propose an end-to-end drone navigation methodology using expert data aided deep reinforcement learning on images acquired by a single camera. The end-to-end approach has been summarized in the block diagram shown in Figure 1. We limit the action space to three actions $\mathcal{A} = \{a_F, a_L, a_R\}$ where under the action $a_F$ the drone moves forward (by $0.25m$), $a_L$ the drone turns left ($45^o$) and $a_R$ the drone turns right ($45^o$). To address the issues of real-time DRL, we explore the following solutions keeping in mind the agent's weight, cost, limited sensing capabilities, and environmental constraints.

### A. Reward generation

Since we are not using any external sensing modalities, the reward needs to be generated from the image frame itself.

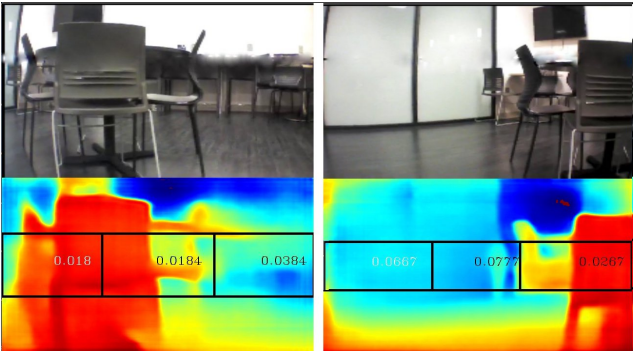

Fig. 2. Depth-based dynamic windowing

**Algorithm 1** Reward generation using the depth map
**function** $f_r(s_t, a_t, s'_t)$
    $d(s_t) \leftarrow$ depth map of $s_t$
    $d(s'_t) \leftarrow$ depth map of $s'_t$
    $d^l(s_t), d^c(s_t), d^r(s_t) = DepthValues(d(s_t))$
    $d^l(s'_t), d^c(s'_t), d^r(s'_t) = DepthValues(d(s'_t))$
    **if** $a_t = a_F$ **then** $r_t = d^c(s'_t)$
    **else if** $a_t = a_L$ **then** $r_t = d^c(s'_t) + \alpha(d^l(s_t) - d^r(s_t))$
    **else**   $r_t = d^c(s'_t) + \alpha(d^r(s_t) - d^l(s_t))$
    **if** $d^c(s'_t) < d_{thresh}$ **then** $r_t = r_{crash}$
    **return** $r_t$

We use the depth map of the state towards the generation of the reward. A depth map of a frame is an image of the same dimension with pixels intensities corresponding to the depth of the pixel in the input image. In the last few years various off-line learning algorithms have been explored to generate depth maps using a single image [15]–[17]. Due to its superior test accuracy, we use the approach proposed in [16].

In order for the reward to be simple and meaningful, we use parts of the depth map towards reward generation. The depth map generated is divided into three windows. The depth in the windows is used to generate a notion of the distance to the closest obstacle in each of the three (left, center and right) directions. This distance is calculated by averaging the smallest $n\%$ pixel depth values. The value of $n$ depends on the nature of the environment. If the environment is expected to have narrow (wide) obstacles, the value of $n$ is relatively smaller (larger). We note that changing the window size dynamically with the global depth in the scene aids reward generation and improves accuracy. If the global depth of the image is greater, then the objects being seen in the frame are farther apart. We choose the relationship between the global depth and window size empirically to be $[H, W]/(0.75 \times global\_depth + 0.5)$ where $[H, W]$ are the dimensions of the input frame from camera. This global depth based dynamic windowing can be seen in Figure 2. The three local distances to the closest obstacle in corresponding directions are then put to use towards reward generation according to Algorithm 1 where $\alpha \in [0, 1]$ is a parametric weight and is taken to be 1/3; $d_{thresh}$ is used to mark the completion of an episode as explained in the next section.

### B. Addressing safety issues

If at any point, the center window shows the distance to the nearest obstacle $d^c$ to be below some threshold value $d_{thresh}$, the agent stops and considers to have "virtually crashed". This virtual crashing marks the end of an episode. Thus the agent does not physically collide with obstacles and significantly reduces the risk of damaging itself or the environment. Once the agent virtually crashes, a penalizing reward $r_{crash}$ is provided to the state-action pair leading to the crash.

### C. Resetting the agent to a suitable initial position

In our approach, the agent does not reset to its initial position, rather a new initial position is selected after the end of every episode. The new initial position is chosen in an autonomous way making use of the knowledge of the "virtual crash" state-action pair. The action that led to the collision is un-done. The agent accomplishes this by taking the opposite actions (for e.g. if the forward action led to virtual crash, the agent after marking it the end of an episode, moves backward) until $d^c$ is at least $d_{recover}$; a threshold set for recovering from the crash. A new episode starts from the recovered state and the policy prevents the agent from selecting the "virtual crash" action for that initial state.

### D. Large online data

*a) Expert Data $D_\mathcal{E}$:* We address the requirement of a large training data set by making the use of Learning from Demonstration (LfD) [18]. At the onset, a human expert navigates the agent across the arena and collects a limited set of expert data-points. The idea of collecting expert data-points is to help the agent through guided exploration. This expert data set is used towards learning in the following two ways.

- Pre-training phase: The neural network is trained for this small set of expert data $D_\mathcal{E}$ and the weights learned $\theta_\mathcal{E}$ are used as initial weights for the online learning phase. This preserves some knowledge about the environment and gives the agent a good starting point for exploration.
- Expert data as a part of experience replay: The expert data is also used as a part of the replay memory $\mathcal{D}_{replay}$ from which the batches of data-points are sampled for training. Making expert data a persistent part of the experience replay helps avoid the neural network from forgetting what it had learned in the pre-training phase.

*b) Knowledge based Data aggregation:* The data aggregation is carried out in the following two ways:

- When the agent virtually crashes, going forward from that state will lead to a crash too. If the agent which is in state $s_t$ moves to the next state $s'_t$ by taking an action $a_t$ and virtually crashes, then the data-point $(s'_t, a_F, s'_t, r_{crash})$ will be aggregated to the current data points.

- Since opposite actions are selected to recover from crashes, the intermediate states will lead to a crash as well. For example, the agent in state $s_t$ moves to next state $s_{t+1}$ by taking an action $a_t$ and virtually crashes. Let $a'_t$ be the opposite action to $a_t$. If $a_t \in \{a_R, a_L\}$ then the data-points $(s_{t+i}, a'_t, s_{t+i-1}, r_{crash})$ and $(s_{t+i}, a_F, s_{t+i}, r_{crash})$ for $i = \{1, 2, 3, ..., n_{recover}\}$ is aggregated to current data-points where $n_{recover}$ is the number of steps required to recover from the crash. Since going backward does not belong to our defined action space $\mathcal{A}$, the data-points are not aggregated if $a_t = a_F$

### E. Convergence of Deep RL algorithm

The basic RL algorithm often suffers from limited convergence, which mainly emerges because the Bellman equation tends to over-estimate the Q-values due to its non-linear nature. Also, the aggregating nature of the Bellman equation might lead to diverging Q-values. So, in order to avoid these issues the following solutions are implemented.

*1) Restricting the range of rewards:* The distances to the nearest obstacle $\{d^l, d^c, d^r\} \in \mathbb{R}^+$ is the estimated distances in meters. These distances are scaled down to have values between $[0, \frac{1}{\alpha+1}]$ where $\alpha$ is the weight constant used in the reward function. When scaled down, the reward function generates the reward within the limited range of $[-1, 1]$

*2) Clipping Q values in Bellman equation:* This ensures that the Q-value updates do not diverge. Let $Q_\theta^{target}(s,a) = r + \gamma \max_a Q(s', a; \theta)$ be the normal Q-value update where $\theta$ is the weights of the neural network, then the clipped Bellman equation is

$$\hat{Q}_\theta^{target}(s,a) = clip(Q_\theta^{target}(s,a), -1, 1) \qquad (2)$$

where the function $clip(a, n_1, n_2)$ clips the value to $n_1$ or $n_2$ if $a$ is less than $n_1$ or greater than $n_2$ respectively. The updated equation ensures that $Q_\theta^{target}(s,a) \in [-1, 1]$ and does not diverge.

*3) Use of Double DQN:* We address the overestimation of the Q value by using a Double Deep Q Network (DDQN) [19]. In DDQN two different copies of neural network are used. One of the neural networks (the behaviour network, $\theta$) is used for training, while the other network (the target network, $\theta'$) is used towards the Bellman equation update. The target network is updated with the weights of the behaviour network after every $n_{target}$ intervals. The updated Bellman equation looks like

$$Q_{\theta'}^{target}(s,a) = r + \gamma \max_{a'} Q(s', a'; \theta') \qquad (3)$$

Combining both clipping and DDQN, the updated Bellman equation is:

$$\hat{Q}_{\theta'}^{target}(s,a) = clip(r + \gamma max_a Q(s', a; \theta'), -1, 1) \qquad (4)$$

### F. Network Architecture

We use a modified AlexNet [20] network to estimate the Q values for the states. The input to the network is the resized camera frame $s_t$. The network consists of 5 convolutional layers and 3 fully-connected layers.

**Algorithm 2** NAVREN-RL Algorithm

Input: Expert data-points: $D_\mathcal{E}$
Initialization: Behaviour network: $Q_\theta(s) = \mathcal{N}(s; \theta)$, Target network: $Q_{\theta'}(s) = \mathcal{N}(s; \theta')$, $m$: Number of pre-training updates, $n_{target}$: Target network update interval, $b_\varepsilon$:$\varepsilon$ annealing coefficient, $n_{batch}$: mini-batch size for training
**for** $i \in \{1, 2, 3, ..., m\}$ **do**
    Sample a mini-batch of size $n_{batch}$ from $D_\mathcal{E}$
    Evaluate the loss $J_{\theta'}(\theta)$
    Perform gradient descent to minimize $J_{\theta'}(\theta)$ w.r.t $\theta$
Initialize the replay memory $\mathcal{D}_{replay} \leftarrow D_\mathcal{E}$
**for** $t \in \{1, 2, 3, ...\}$ **do**
    $s_t \leftarrow$ Camera image, $Q(s_t) \leftarrow \mathcal{N}(s_t; \theta)$
    Sample an action from behaviour policy $a_t \sim \pi^{b_\varepsilon Q}(\varepsilon)$
    Implement the action $a_t$ on the agent
    $s'_t \leftarrow$ Camera image, $Q(s'_t) \leftarrow \mathcal{N}(s'_t; \theta)$
    Generate the reward $r_t \leftarrow f_r(s_t, a_t, s'_t)$
    Store the tuple $(s_t, a_t, s'_t, r_t)$ in $\mathcal{D}_{replay}$
    **if** virtual crash **then**
        **while** not recover from crash **do**
            Aggregate data-points to $\mathcal{D}_{replay}$
    Sample a mini-batch of size $n_{batch}$ from $\mathcal{D}_{replay}$
    Evaluate the loss $J_{\theta'}(\theta)$
    Perform gradient descent to minimize $J_{\theta'}(\theta)$ w.r.t $\theta$
    **if** $t$ mod $n_{target} = 0$ **then** $\theta \leftarrow \theta'$

### G. Online Learning

Before the learning process begins, an expert user navigates the agent in the selected environment for a certain number of steps $n_{expert}$. The data tuple $(s_i, a_i, s'_i, r_i)$ for each of the steps $i \in \{1, 2, 3, ..., n_{expert}\}$ is generated and saved in $D_\mathcal{E}$. Next comes the pre-training phase where random mini-batches of size $n_{batch}$ are selected from the expert data $D_\mathcal{E}$ and a neural network $Q_\theta(s) = \mathcal{N}(s; \theta)$ is trained minimizing

$$J_{\theta'}(\theta) = \sum_{i=1}^{n_{batch}} J(s_i, a_i, \theta, \theta') + \beta J_{reg}(\theta) \qquad (5)$$

where $J(s_i, a_i, \theta, \theta')$ is the TD loss for $i^{th}$ data-point dictated by the Bellman equation and $J_{reg}(\theta)$ is regularization loss to help prevent over-fitting the network for the smaller amount of expert data, and $\beta$ is a regularization weight. These losses are given by:

$$J(s_i, a_i, \theta, \theta') = ||\hat{Q}_{\theta'}^{target}(s_i, a_i) - Q(s_i, a_i; \theta)||_2 \qquad (6)$$

$$J_{reg}(\theta) = ||\theta||_2 \qquad (7)$$

where $\hat{Q}_{\theta'}^{target}(s_t, a_t)$ is given by equation 4

After the pre-training phase, the online training phase begins. The agent is placed in the environment, and follows a $\varepsilon$-greedy policy for actions. with $b_\varepsilon$ as the annealing coefficient. $\varepsilon$ is varied linearly from 0.1 to 0.9 as the number of data-points varies from 1 to $b_\varepsilon$. At every time step $t$, the drone saves the data points $(s_t, a_t, s'_t, r_t)$ in $\mathcal{D}_{replay}$. A mini-batch of size $n_{batch}$ is randomly sampled from the replay memory $\mathcal{D}_{replay}$ and used to minimize the loss defined in equation 5 through gradient descent. Algorithm 2 shows the complete algorithm, while table I lists the hyper parameters used.

TABLE I
LIST OF HYPER PARAMETERS USED FOR TRAINING

| Learning rate | 1e-6 | $n_{target}$ | 200 | $n_{batch}$ | 32 |
|---|---|---|---|---|---|
| $\beta$ | 0.001 | $d_{thresh}$ | 0.02 | $r_{crash}$ | -1 |

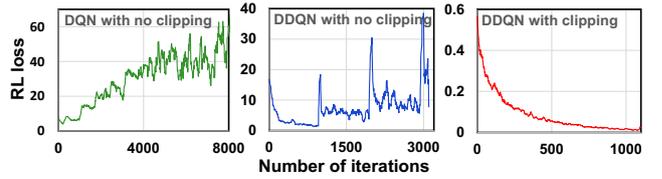

Fig. 4. Convergence of RL with and without DDQN and clipping

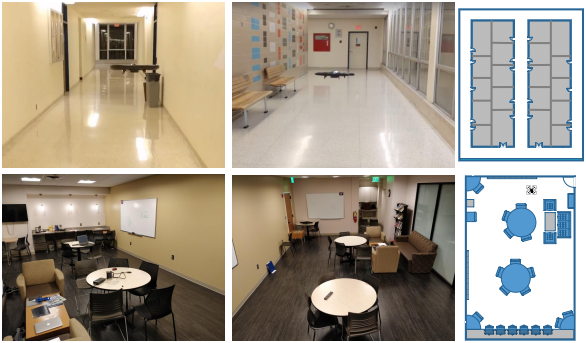

Fig. 3. Snapshots and the layouts of the arenas used. Top row: $A_1$ Hallway, Bottom row: $A_2$ SC-room

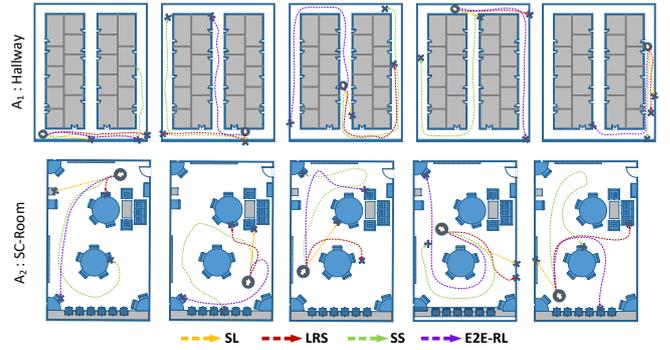

Fig. 5. Trajectories followed by the baselines and NAVREN-RL for 5 different initial locations

## V. EXPERIMENTAL RESULTS

Real-time experimentation are carried out to validate the proposed approach for drone navigation.

### A. Hardware specifications

We use a low cost Parrot AR drone 2.0 which does not have the computational power to carry out the required processing on-board. Hence, the drone sends the camera frames to a workstation/cloud equipped with a core i7 processor and GTX1080 GPUs. Control actions are communicated back to the drone. We use Tensorflow to carry out the neural network computation on the workstation.

### B. Testing environments

We use the following two arenas to carry out the experimentation for successful navigation.

*1) Arena $A_1$: Open Hallway:* This is a hallway in an engineering building with glass walls. The drone has to navigate through the narrow hallways (minimum width of $\approx 1.5m$). There are no extra obstacles between the hallway path except for water dispenser, benches and trashcans.

*2) Arena $A_2$: SC Room:* This arena is a cluttered break-out room with couches, chairs, tables and bar-stools with narrow passages in between ($\approx 1m$).

The layout and floor plans of these arenas can be seen in the Figure 3

### C. Baseline Algorithms for Comparison

We compare our method with the following baseline algorithms.

*1) Straight-line controller:* This controller always predicts the forward action hence moving the agent in a straight line in a manner described in [14]. This controller provides a good comparison of the complexity of the arena.

*2) Left-Right-Straight (LRS) controller:* This baseline is based on the work in [13] where a supervised approach is used to classify images with respect to the actions required to be taken. A human expert roams around the arena and collects the images using left, right and forward facing cameras. Images collected from left (right) facing camera are labeled with the target action of right (left) while the ones collected from forward facing camera are labeled with the target action of forward. These labeled images are then used to train a neural network offline in a supervised manner.

*3) Self-supervised (SS) controller:* This controller uses the work proposed in [10] where a large data-set of crash and safe images are collected over various indoor environments. These labeled images are then used to train a neural network to classify each image as either safe or crash. In the inference phase, the input camera frame is then divided into three windows and the probability of crash in each of the sub-frames is calculated. Based on these probabilities, a handcrafted controller is designed, following [10] to take suitable actions.

### D. Performance

Figure 4 shows the comparison of RL convergence with and without DDQN and clipping of the Bellman equation. It can be seen that the DDQN with clipping of Bellman equation shows good convergence.

We assess the performance of NAVREN-RL by comparing it against the baselines mentioned above. 3000 (700 expert+2300

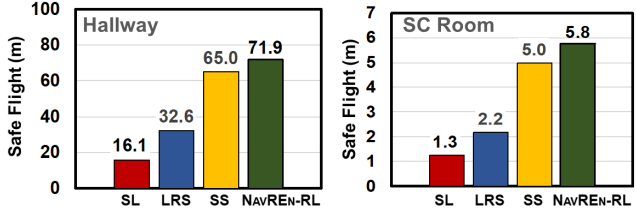

Fig. 6. Safe flight (in meters) comparison across baselines

TABLE II
ARENA STATS

| Arena | Method | Total Distance(m) | Safe Flight(m) | Improvement |
|---|---|---|---|---|
| Hallway | SL | 80.7 | 16.1 | 4.45x |
| | LRS [13] | 162.9 | 32.6 | 2.21x |
| | SS [10] | 324.9 | 65.0 | 1.11x |
| | NAVREN-RL (ours) | 359.5 | 71.9 | – |
| SC-room | SL | 6.3 | 1.3 | 4.55x |
| | LRS [13] | 10.9 | 2.2 | 2.65x |
| | SS [10] | 24.9 | 5.0 | 1.16 x |
| | NAVREN-RL (ours) | 28.8 | 5.8 | – |

online) data-points are collected in the Hallway arena, while 2000 (600 expert+1400 online) data-points are collected in the SC-room arena. In each of the arenas, all the 4 techniques are separately used to learn a neural network. The agent is initialized by the learned neural network and the performance is evaluated by placing the drone at 5 different initial locations. To keep the comparison fair, the agent is placed precisely the same way across all the techniques. In each of the cases, the trajectory followed by the agent is recorded until the agent is no longer able to navigate. This loss of navigation is considered if

- The agent collides into an obstacle
- The agent keeps hovering, being stuck in a repetitive pattern of left/right actions, and does not move forward for 10 iterations

The trajectories can be seen in Figure 5. The distance covered by the agent before crash is taken to be the performance metric and can be seen in the table II. The total distance covered is the sum of the individual distances covered by the drone from each of the initial locations. The safe flight for any technique is the average distance covered by the drone from the different initial locations. In most of the cases the proposed NAVREN-RL method outperforms the baselines, i.e the safe flight (m) for the proposed RL method is the highest among the baselines. This can be seen in Figure 6.

## VI. CONCLUSIONS

This paper provides an end-to-end reinforcement learning algorithm for autonomous navigation of drones in indoor real environments by addressing the problems associated with the RL implementation. Experimentation is carried out in different arenas and the performance is compared to other base-lines. The results show that the agent is able to navigate in the indoor arena with limited sensing capabilities and data-points with comparable performance.

ACKNOWLEDGMENTS

This work was supported in part by C-BRIC, one of six centers in JUMP, a Semiconductor Research Corporation (SRC) program sponsored by DARPA.

REFERENCES

[1] Y. K. Kwag and J. W. Kang, "Obstacle awareness and collision avoidance radar sensor system for low-altitude flying smart uav," in Digital Avionics Systems Conference, 2004. DASC 04. The 23rd, vol. 2. IEEE, 2004, pp. 12–D.
[2] A. S. L. Raimundo et al., "Autonomous obstacle collision avoidance system for uavs in rescue operations," Ph.D. dissertation, 2016.
[3] D. Silver, A. Huang, C. J. Maddison, A. Guez, L. Sifre, G. Van Den Driessche, J. Schrittwieser, I. Antonoglou, V. Panneershelvam, M. Lanctot et al., "Mastering the game of go with deep neural networks and tree search," nature, vol. 529, no. 7587, p. 484, 2016.
[4] V. Mnih, K. Kavukcuoglu, D. Silver, A. A. Rusu, J. Veness, M. G. Bellemare, A. Graves, M. Riedmiller, A. K. Fidjeland, G. Ostrovski et al., "Human-level control through deep reinforcement learning," Nature, vol. 518, no. 7540, p. 529, 2015.
[5] D. O. Sales, P. Shinzato, G. Pessin, D. F. Wolf, and F. S. Osorio, "Vision-based autonomous navigation system using ann and fsm control," in Robotics Symposium and Intelligent Robotic Meeting (LARS), 2010 Latin American. IEEE, 2010, pp. 85–90.
[6] R. Huang, P. Tan, and B. M. Chen, "Monocular vision-based autonomous navigation system on a toy quadcopter in unknown environments," in Unmanned Aircraft Systems (ICUAS), 2015 International Conference on. IEEE, 2015, pp. 1260–1269.
[7] C. D. Schuman, T. E. Potok, R. M. Patton, J. D. Birdwell, M. E. Dean, G. S. Rose, and J. S. Plank, "A survey of neuromorphic computing and neural networks in hardware," arXiv preprint arXiv:1705.06963, 2017.
[8] C. Richter and N. Roy, "Safe visual navigation via deep learning and novelty detection," in Proc. of the Robotics: Science and Systems Conference, 2017.
[9] L. Tai, S. Li, and M. Liu, "Autonomous exploration of mobile robots through deep neural networks," International Journal of Advanced Robotic Systems, vol. 14, no. 4, p. 1729881417703571, 2017.
[10] D. Gandhi, L. Pinto, and A. Gupta, "Learning to fly by crashing," arXiv preprint arXiv:1704.05588, 2017.
[11] D. K. Kim and T. Chen, "Deep neural network for real-time autonomous indoor navigation," arXiv preprint arXiv:1511.04668, 2015.
[12] A. Giusti, J. Guzzi, D. C. Cireşan, F.-L. He, J. P. Rodríguez, F. Fontana, M. Faessler, C. Forster, J. Schmidhuber, G. Di Caro et al., "A machine learning approach to visual perception of forest trails for mobile robots," IEEE Robotics and Automation Letters, vol. 1, no. 2, pp. 661–667, 2016.
[13] S. Ross, N. Melik-Barkhudarov, K. S. Shankar, A. Wendel, D. Dey, J. A. Bagnell, and M. Hebert, "Learning monocular reactive uav control in cluttered natural environments," in Robotics and Automation (ICRA), 2013 IEEE International Conference on. IEEE, 2013, pp. 1765–1772.
[14] F. Sadeghi and S. Levine, "Cad2rl: Real single-image flight without a single real image," arXiv preprint arXiv:1611.04201, 2016.
[15] A. Saxena, S. H. Chung, and A. Y. Ng, "3-d depth reconstruction from a single still image," International journal of computer vision, vol. 76, no. 1, pp. 53–69, 2008.
[16] I. Laina, C. Rupprecht, V. Belagiannis, F. Tombari, and N. Navab, "Deeper depth prediction with fully convolutional residual networks," in 3D Vision (3DV), 2016 Fourth International Conference on. IEEE, 2016, pp. 239–248.
[17] C. Godard, O. Mac Aodha, and G. J. Brostow, "Unsupervised monocular depth estimation with left-right consistency," in CVPR, vol. 2, no. 6, 2017, p. 7.
[18] T. Hester, M. Vecerik, O. Pietquin, M. Lanctot, T. Schaul, B. Piot, D. Horgan, J. Quan, A. Sendonaris, G. Dulac-Arnold et al., "Deep q-learning from demonstrations," arXiv preprint arXiv:1704.03732, 2017.
[19] H. Van Hasselt, A. Guez, and D. Silver, "Deep reinforcement learning with double q-learning." in AAAI, vol. 2. Phoenix, AZ, 2016, p. 5.
[20] A. Krizhevsky, I. Sutskever, and G. E. Hinton, "Imagenet classification with deep convolutional neural networks," in Advances in neural information processing systems, 2012, pp. 1097–1105.